\documentclass[letterpaper, 10 pt, conference]{ieeeconf}  

\IEEEoverridecommandlockouts                              

\usepackage{url}
\usepackage{caption}
\usepackage{subcaption}
\usepackage{graphicx}
\usepackage{cite}
\usepackage{makecell}
\usepackage{multirow}
\usepackage{enumerate}
\usepackage{amsmath}
\usepackage{etoolbox}
\usepackage{placeins}
\usepackage{hyperref}
\usepackage{setspace} 
\usepackage{romannum}
\usepackage{wasysym}

\usepackage{bbding}
\usepackage{pifont}
\usepackage{amssymb}
\usepackage{xcolor}
\usepackage[outline]{contour}
\usepackage{cuted} 

\DeclareMathAlphabet\mathbfcal{OMS}{cmsy}{b}{n}

\title{\LARGE \bf ManipForce: Force-Guided Policy Learning with \\ Frequency-Aware Representation for Contact-Rich Manipulation}

\author{%
Geonhyup Lee$^{1}$,
Yeongjin Lee$^{1}$,
Kangmin Kim$^{1}$,
Seongju Lee$^{1}$,
Sangjun Noh$^{1}$,
Seunghyeok Back$^{2}$,
Kyoobin Lee$^{1\dagger}$%
\thanks{$^{1}$ G. Lee, Y. Lee, K. Kim, S. Lee, S. Noh, and K. Lee are with the Department of AI Convergence, Gwangju Institute of Science and Technology (GIST), Gwangju 61005, Republic of Korea.}%
\thanks{$^{2}$ S. Back is with the Department of AI Machinery, Korea Institute of Machinery \& Materials (KIMM), Daejeon 34103, Republic of Korea.}%
\thanks{$^{\dagger}$ Corresponding author: Kyoobin Lee {\tt\small kyoobinlee@gist.ac.kr}}%
}

\begin{document}
\maketitle
\begin{strip}
\vspace{-2.9cm} 

\centering
\includegraphics[width=0.98\textwidth]{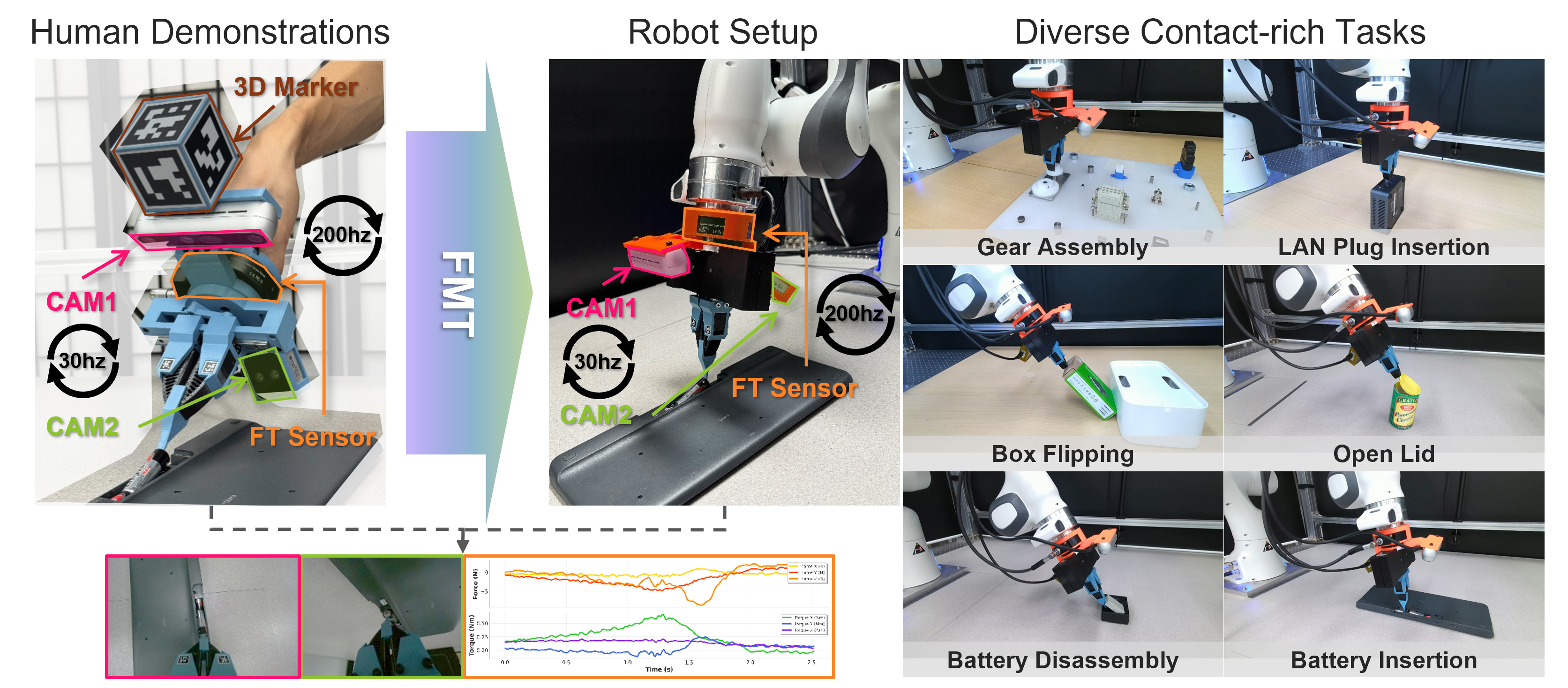}
\captionof{figure}{Overview of our framework.
(Left) \textbf{ManipForce}: a handheld system with dual cameras and a wrist-mounted F/T sensor capturing RGB–F/T data from human demonstrations.
(Middle) The same configuration on a robot enables direct transfer to real-world execution.
(Right) Six contact-rich tasks—gear assembly, LAN plug insertion, box flipping, open lid, battery disassembly, and battery insertion—used to evaluate our \textbf{FMT} model.}
\label{fig:intro_examples}
\end{strip}

\begin{abstract}
Contact-rich manipulation tasks such as precision assembly require precise control of interaction forces, yet existing imitation learning methods rely mainly on vision-only demonstrations. We propose ManipForce, a handheld system designed to capture high-frequency force–torque (F/T) and RGB data during natural human demonstrations for contact-rich manipulation. Building on these demonstrations, we introduce the Frequency-Aware Multimodal Transformer (FMT). FMT encodes asynchronous RGB and F/T signals using frequency- and modality-aware embeddings and fuses them via bi-directional cross-attention within a transformer diffusion policy. Through extensive experiments on six real-world contact-rich manipulation tasks—such as gear assembly, box flipping, and battery insertion—FMT trained on ManipForce demonstrations achieves robust performance with an average success rate of 83\% across all tasks, substantially outperforming RGB-only baselines. Ablation and sampling-frequency analyses further confirm that incorporating high-frequency F/T data and cross-modal integration improves policy performance, especially in tasks demanding high precision and stable contact. 
Hardware, software, and video demos are available at: \url{https://sites.google.com/view/manipforce/홈}.
\end{abstract}

\section{Introduction}
	
Contact-rich manipulation tasks such as precise assembly~\cite{luo2019reinforcement, zhang2024residual, chen2023multimodality, lee2024polyfit}, battery disassembly~\cite{kang2025robotic}, and non-prehensile handling~\cite{hou2024adaptive} require high precision and \textbf{force-aware manipulation}. Humans naturally perceive contact forces and their subtle changes when assembling parts, adjusting their strategies accordingly. Yet most robotic approaches rely solely on visual demonstrations, missing the rich F/T information humans provide.

Recent advances in imitation learning~\cite{chi2023diffusion, ze20243d, zhao2023learning} have demonstrated strong potential for dexterous and contact-rich manipulation by learning directly from human demonstrations.
However, these methods still rely on high-quality demonstration data, which is costly and difficult to collect for fine-grained interactions.
Hand-held data collection systems such as UMI~\cite{chi2024universal} have been proposed to address this challenge by enabling natural human demonstrations without the expertise requirements and remote-control limitations of teleoperation.
While effective for simplifying demonstration collection, UMI does not capture force–torque (F/T) information, which is essential for accurately modeling contact behaviors.
More recent work~\cite{liu2024forcemimic} combines visual and F/T data but relies on point clouds to represent the scene, which introduces complex setup requirements and fundamentally limits the ability to perceive small objects and fine clearances essential for contact-rich manipulation.
Furthermore, from a learning perspective, this approach down-samples high-frequency F/T signals to match the image frame rate, losing rich temporal information necessary for modeling contact dynamics.

To address these limitations, we introduce \textbf{ManipForce} a handheld system for simultaneous RGB–F/T data collection during natural human demonstrations, and the \textbf{Frequency-Aware Multimodal Transformer (FMT)}, which learns robust policies from the collected data for diverse, precise, and contact-rich manipulation tasks.

\textbf{ManipForce} consists of a dual handheld camera setup with a wrist-mounted F/T sensor to capture both visual and high-frequency force signals during human-guided demonstrations.
This configuration enables robust perception of small objects, tight clearances, and fine-grained contacts, allowing collected demonstrations to transfer directly to robotic execution.
We replace SLAM-based wrist tracking with 3D ArUco marker pose estimation to maintain accuracy during close-contact interactions without environmental dependencies, and apply tool gravity compensation to ensure precise and interaction-focused F/T measurements.
We propose the \textbf{FMT}, which learns from asynchronous RGB (30 Hz) and F/T (\textgreater200 Hz) signals using a Transformer-based Diffusion Policy~\cite{chi2023diffusion} architecture.
To exploit the higher-frequency force signals relative to images, the model tokenizes both RGB and F/T inputs using learnable frequency and modality embeddings.
This design enables the model to effectively handle heterogeneous modalities with asynchronous sampling rates.
In addition, bi-directional cross-attention modules fuse complementary information across modalities.
We evaluate our approach on six contact-rich manipulation tasks spanning precision assembly, non-prehensile manipulation, and complex disassembly, and observe significant performance gains over RGB-only baselines.
Ablation studies further confirm that high-frequency F/T sensing, unified positional embeddings, and bi-directional cross-attention each make complementary contributions to robust multimodal policy learning.

Our main contributions are:
\begin{itemize}
\item We introduce \textbf{ManipForce}, a handheld RGB–F/T data collection system enabling diverse and fine-grained contact-rich manipulation demonstrations.
\item We propose \textbf{FMT}, which handles inputs with asynchronous sampling rates through frequency-aware multimodal representation learning and cross-attention within a Transformer architecture, enabling robust policy learning for contact-rich manipulation.
\item We demonstrate robust performance on diverse contact-rich manipulation tasks—including gear assembly, plug insertion, battery disassembly, and lid operations—consistently outperforming RGB-only baselines.
\end{itemize}

\section{Related Work}
\subsection{Human Demonstration Data Collection}

Teleoperation-based human demonstration collection includes VR-based systems~\cite{iyer2024open, ding2024bunny, cheng2024open}, 3D spacemouse controllers~\cite{luo2025fmb, ankile2024juicer}, and leader–follower systems with joint mapping~\cite{zhao2023learning, wu2024gello, fang2024airexo, yang2024ace, huang2025dih, liu2025factr, myers2025child}. While leader–follower systems provide intuitive control, teleoperation takes more time, needs skilled operators, and lacks clear visual and haptic feedback. Hand-held data collection systems, such as UMI\cite{chi2024universal}, have shown a promising direction, enabling intuitive data collection by allowing users to directly manipulate a hand-held gripper. This paradigm has since expanded to include additional sensing modalities—audio~\cite{liu2024maniwav}, tactile~\cite{liu2025vitamin, zhu2025touch, wu2025freetacman, huang2025tactile}, and F/T~\cite{liu2024forcemimic, adeniji2025feel, chen2025dexforce}. However, current F/T data collection approaches remain limited. Specialized fingertip force sensors and fixed-wrist setups are required~\cite{chen2025dexforce}. Others~\cite{liu2024forcemimic} combine visual and F/T data but rely on manual cropping for data preparation, depth-based point clouds for scene representation, and visual-SLAM tracking for pose estimation. These steps introduce complex setup requirements, make it difficult to capture small features or precise depth in tasks such as LAN plug insertion, and reduce accuracy during close-contact interactions. Our proposed system addresses these limitations by enabling direct RGB–F/T data collection with a hand-eye camera setup, supporting diverse and fine-grained manipulation tasks without task-specific calibration, and employing marker-based pose tracking to maintain accuracy during close-contact interactions.

\begin{figure*}[t]
    \centering
    \includegraphics[width=1\linewidth]{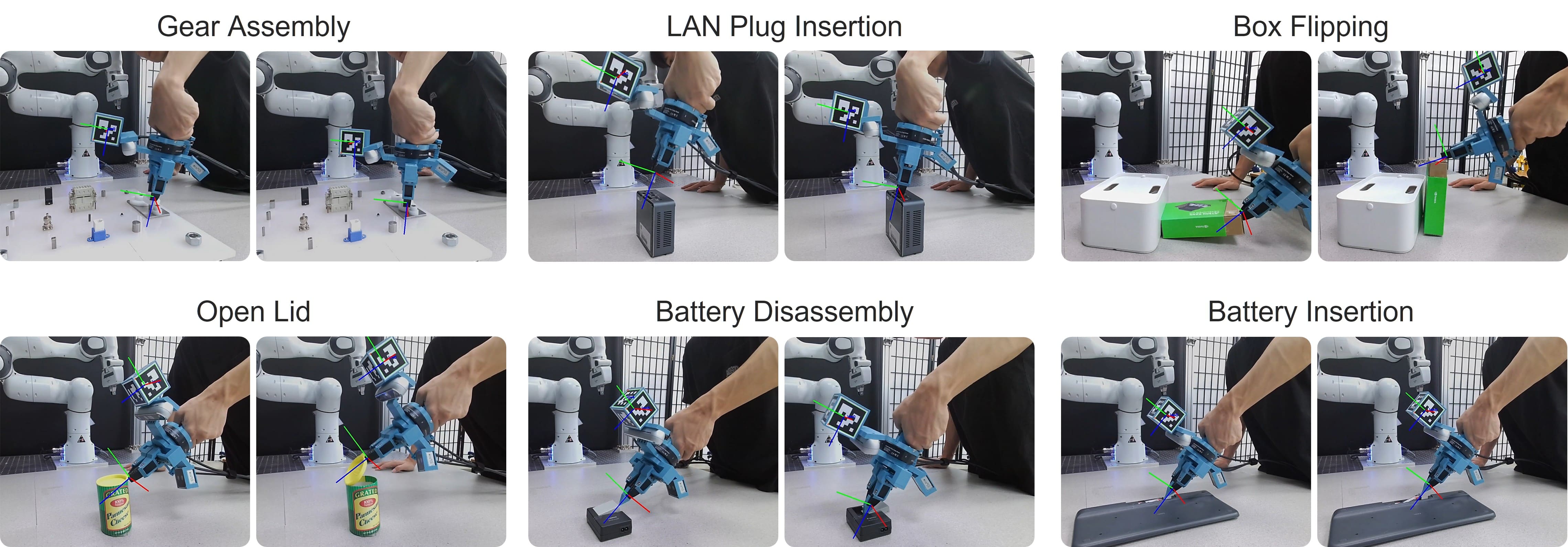}
    \caption{Six contact-rich manipulation tasks used in our evaluation: Gear Assembly, LAN Plug Insertion, Box Flipping, Open Lid, Battery Disassembly, and Battery Insertion. These tasks require precise force control and multimodal feedback, testing our model’s ability to integrate high-frequency force signals with visual input for robust manipulation.}
    \label{fig:Evaluation_Task}
\end{figure*}

\subsection{Multimodal Imitation Learning}

Recent advances in visual imitation learning~\cite{chi2023diffusion, ze20243d} have shown robustness by modeling complex action distributions directly from images. However, vision-only approaches omit F/T cues that are critical for contact-rich manipulation. 
To address this limitation, multimodal approaches combine vision with force sensing through various strategies, including feature-level concatenation~\cite{liu2024forcemimic, yang2023moma, liu2025factr}, and dynamic modality weighting via contact prediction~\cite{he2024foar}. Other studies introduce cross-attention modules for vision–F/T integration~\cite{kang2025robotic}, or adaptive compliance control with learned dynamic gains~\cite{hou2024adaptive}. 
Parallel work on vision–tactile integration tackles similar challenges through feature concatenation~\cite{huang20243d, luu2025manifeel}, force-guided cross-attention~\cite{li2025adaptive}, and CLIP-based visual–tactile representation learning~\cite{liu2025vitamin}. A key challenge across these multimodal settings is managing asynchronous, rate-mismatched inputs, where visual sensors operate at low frequencies while force or tactile sensors operate at much higher rates. Although recent efforts such as Reactive Diffusion Policy~\cite{xue2024reactive} introduce dual-architecture designs to address this issue, they still impose complex structures and restrict cross-modal interaction to low-frequency visual streams. We address these limitations with FMT, a Transformer-based Diffusion Policy that uses learnable modality and frequency embeddings together with bi-directional cross-modal attention, enabling frequency-aware multimodal representation learning across asynchronous multimodal inputs.

\section{Method}

In this section, we present our two-component framework combining data collection and policy learning. 
\textbf{ManipForce} (Section~\ref{subsec:data_collection_system}) is a handheld RGB–F/T system that captures high-precision poses and high-bandwidth force signals during natural human demonstrations, yielding diverse and fine-grained data. 
Building on this data, the \textbf{FMT} (Section~\ref{subsec:policy}) uses frequency–aware multimodal embeddings and bi-directional cross-attention to learn robust policies from asynchronous visual and force inputs.

\subsection{ManipForce}
\label{subsec:data_collection_system}

\begin{figure}[t]
    \centering
    \includegraphics[width=0.93\linewidth]{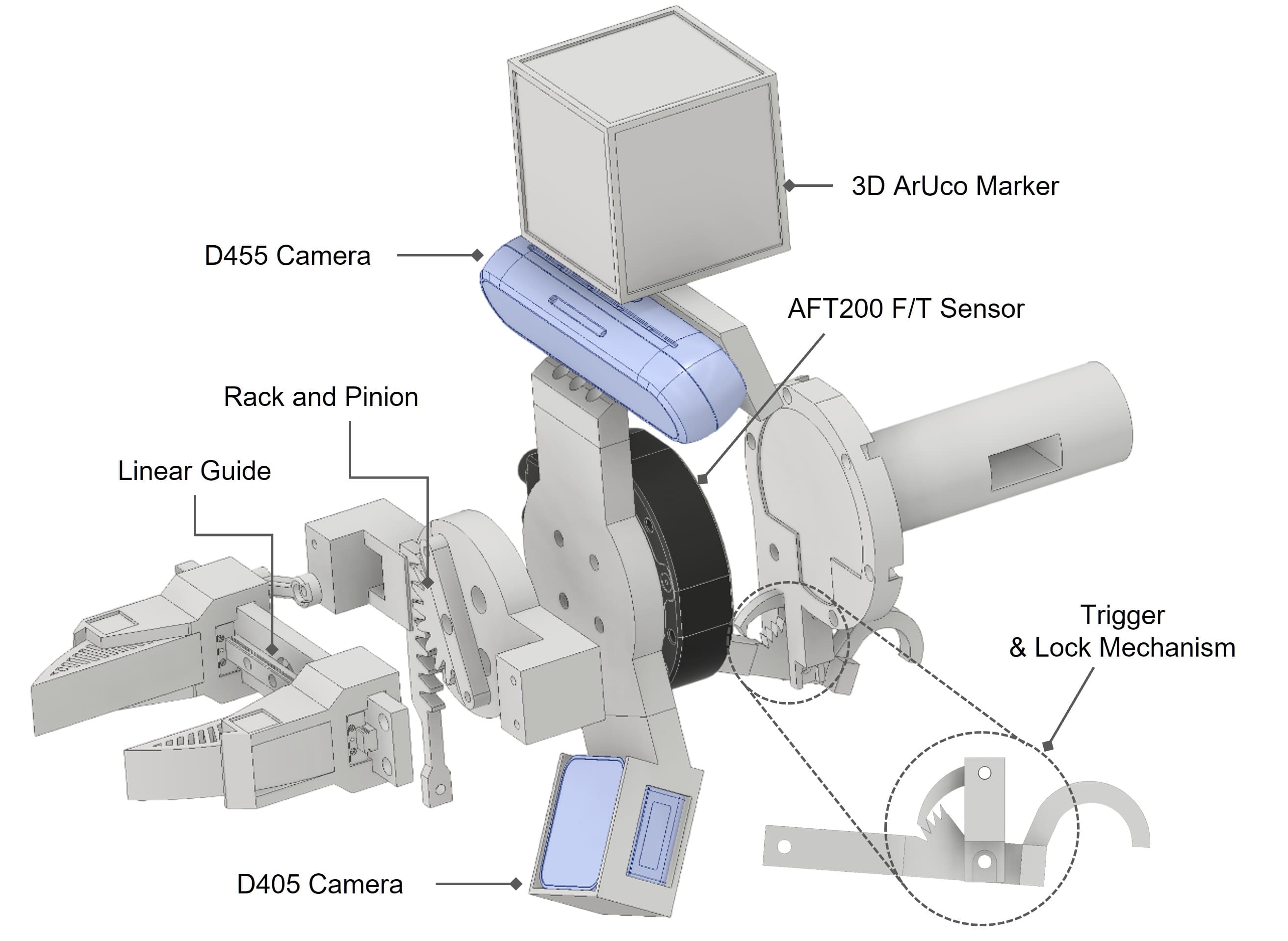}
    \caption{Structure of the ManipForce. It includes dual RGB cameras, a wrist-mounted F/T sensor, and ArUco marker tracking. A rack-and-pinion gripper with trigger-lock and linear guides enables precise, stable human demonstrations.}
    \label{fig:CAD}
\end{figure}

\begin{figure}[t]
    \centering
    \includegraphics[width=1.01\linewidth]{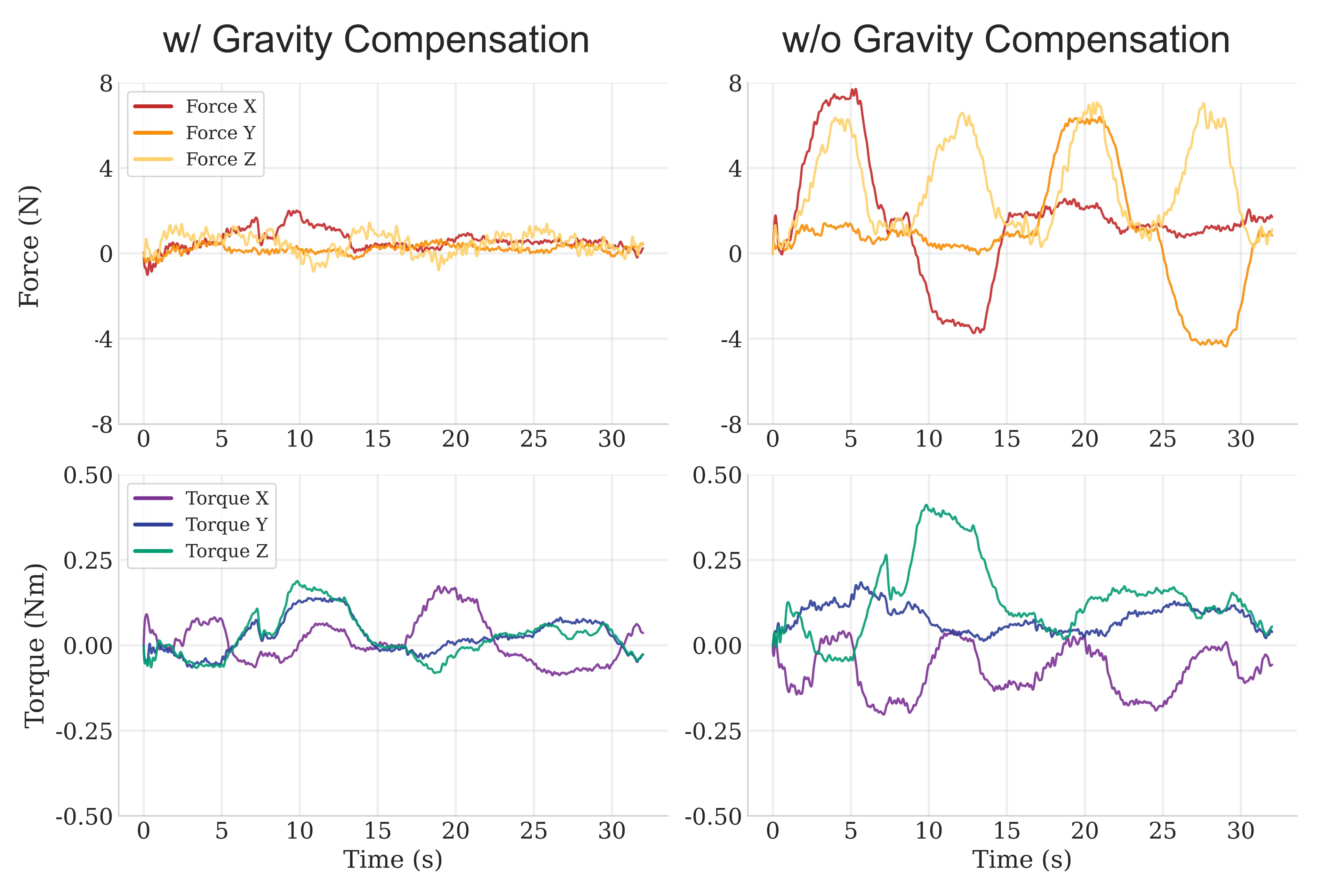}
    \caption{Gravity compensation result in ManipForce. Tilting the system by $\pm90^\circ$ about the X and Y axes shows that the measured F/T signals remain stable despite changes in orientation.}
    \label{fig:gravity_comp}
\end{figure}

\textbf{System Design.} 
Our gripper mechanism is purpose-built to make human-guided data collection intuitive, precise, and stable.
As shown in Fig.~\ref{fig:CAD}, the design integrates several complementary features that improve usability and data quality.
At its core, a rack-and-pinion mechanism with a familiar trigger interface drives parallel jaw motion, allowing users to perform natural and accurate grasping actions.
A deformable pin-ray structure at the fingertips increases compliance and contact sensitivity, while integrated linear guides at the gripper attachment points suppress mechanical vibration and promote smoother, more precise manipulation.
A trigger-lock mechanism holds the jaws closed after grasping, reducing user fatigue and ensuring that recorded signals primarily reflect task-relevant interaction forces.
Finally, all gripper components are mounted downstream of the F/T sensor to capture the complete interaction forces during operation. All hardware and software components are released as open source at our project website.

\textbf{RGB and Wrist-Pose Data Acquisition.} 
The system employs two hand-eye cameras (Intel RealSense D455 and D405) positioned to minimize occlusions and ensure clear visibility during precision manipulation tasks. 
Conventional SLAM-based wrist-pose estimation degrades near surfaces and cannot meet the precision required for fine assembly tasks such as gear or LAN plug insertion. 
To address this, we use 3D ArUco markers detected by an externally positioned Azure Kinect sensor, achieving robust, viewpoint-independent tracking with sub-millimeter accuracy~\cite{garrido2014automatic}. 
End-effector actions are computed as 6-DOF pose deltas between consecutive frames and transformed from the marker reference frame to the robot’s TCP frame using CAD-derived transformation matrices. 
Additional ArUco markers attached to both gripper jaws provide real-time tracking of jaw positions. 
The velocity of these markers between consecutive frames is further analyzed to infer gripper open/close states, enabling detailed annotation of manipulation events. 
All RGB and wrist-pose data are recorded at 30~Hz.

\begin{figure*}[t]
    \centering
    \includegraphics[width=0.95\textwidth]{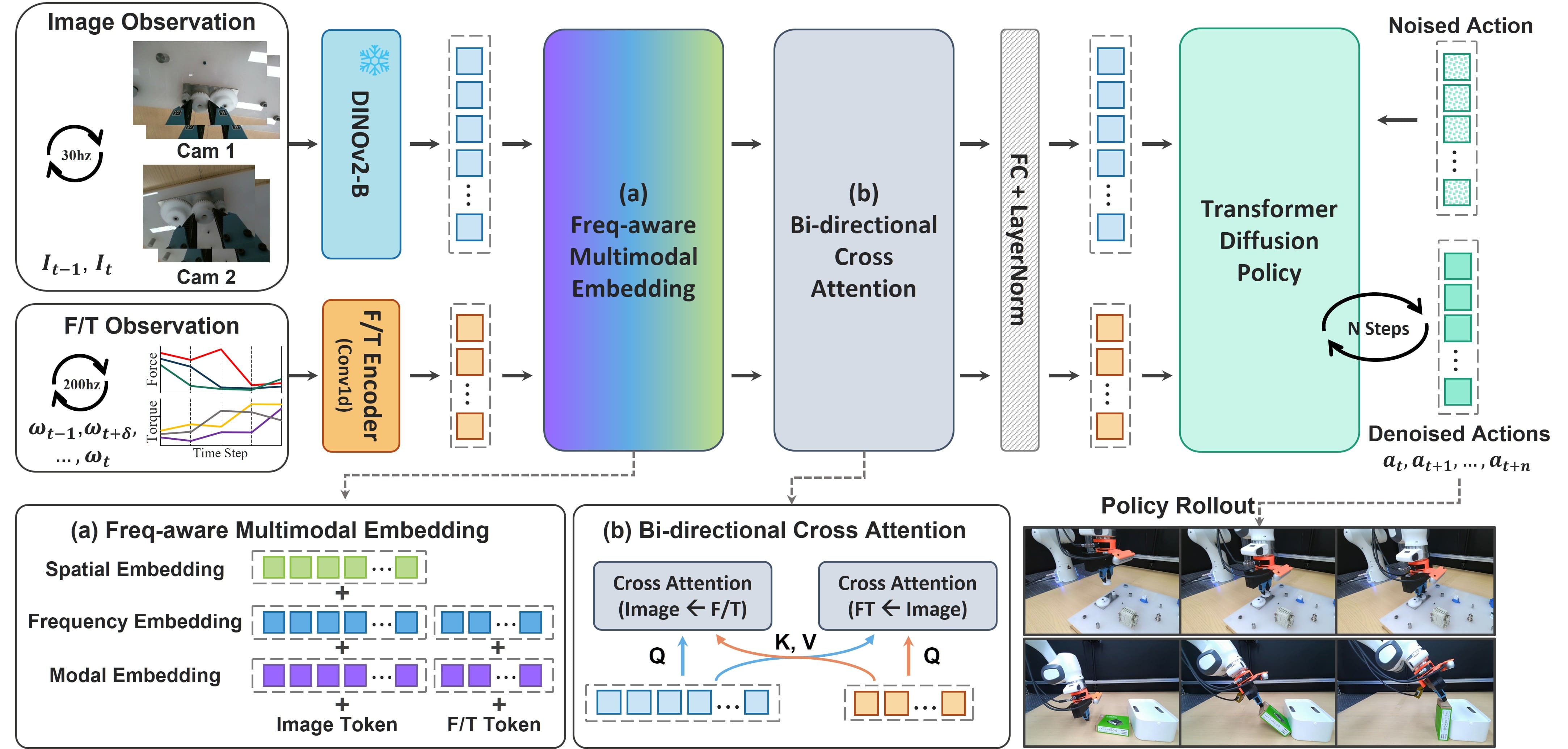}
    \caption{Architecture of FMT. Asynchronous RGB (30~Hz) and F/T (200~Hz) signals are encoded with frequency-aware multimodal embeddings, fused via bi-directional cross-attention, and passed through a Transformer Diffusion Policy to produce robust actions for precise, contact-rich manipulation tasks.}
    \label{fig:model}
\end{figure*}

\textbf{F/T Data Acquisition.} 
An AIDIN AFT200 F/T sensor mounted on the wrist records force/torque signals at 200~Hz.
Because raw F/T measurements include both interaction forces and the weight of the tool, all F/T signals are gravity-compensated using IMU data from the D455 camera to isolate true contact forces and ensure consistency between the ManipForce system and robot platforms. 
The gravity vector from the IMU frame is transformed into the F/T sensor frame as
\[
\mathbf{g}_{ft} = R^{ft}_{imu} \, \mathbf{g}_{imu},
\]
where \(R^{ft}_{imu}\) denotes the rotation from the IMU to the F/T sensor frame. 
The gravitational wrench is then computed as
\[
\mathbf{F}_{g} = m_{tool} \, \mathbf{g}_{ft}, 
\quad
\boldsymbol{\tau}_{g} = \mathbf{r}_{com} \times \mathbf{F}_{g},
\]
and subtracted from the measured wrench to obtain the compensated values:
\[
\mathbf{F}_{comp} = \mathbf{F}_{measure} - \mathbf{F}_{g}, 
\quad
\boldsymbol{\tau}_{comp} = \boldsymbol{\tau}_{measure} - \boldsymbol{\tau}_{g}.
\]
This gravity compensation procedure is applied identically during both handheld data collection and robot execution to ensure consistent F/T measurements across systems. To validate the effectiveness of this compensation, the handheld system was rotated about the x- and y-axes by ±90° while measuring the F/T signals (Fig.~\ref{fig:gravity_comp}). Without compensation, force readings reached up to ±8N and torques up to 0.4N·m, whereas with compensation, the residual forces were reduced to within ±1N and torques to within 0.15~N·m, confirming that the procedure effectively removes gravity-induced bias.

\subsection{Frequency-aware Multimodal Transformer}
\label{subsec:policy}
In this section, we present \textbf{FMT}, a model that learns frequency-aware multimodal representations to fuse low-rate RGB with high-frequency F/T data for contact-rich manipulation. Extending Diffusion Policy~\cite{chi2023diffusion}, FMT applies frequency-aware multimodal embeddings and bi-directional cross-attention to align visual cues with fine-grained force information. We outline its four components—multimodal tokenization, frequency–modality embeddings, cross-attention fusion, and the diffusion-based policy head—that together enable robust learning from asynchronous multimodal inputs. The overall architecture of FMT is illustrated in Fig.~\ref{fig:model}.

\textbf{RGB-Force Tokenization.} Visual observations from dual hand-eye cameras are preprocessed through squared padding and resizing 256×256 resolution before being encoded using DINOv2-B \cite{oquab2023dinov2} to extract visual representations, while F/T representation is extracted via a 1D CNN encoder to capture force dynamics. Specifically, the visual representations from the two hand-eye cameras are denoted by $\mathbf{T}_{\mathrm{cam1}}, \mathbf{T}_{\mathrm{cam2}}\in\mathbb{R}^{T_{\mathrm{img}}\times L\times d}$, and the F/T representation by $\mathbf{T}_{\mathrm{ft}}\in\mathbb{R}^{T_{\mathrm{ft}}\times d}$. Here, $T_{\mathrm{img}}$ is the time horizon of the hand-eye camera, $L$ is the number of visual tokens within an image, $d$ is the model dimension, and $T_{\mathrm{ft}}$ is the time horizon of the F/T observations. We set $T_{\mathrm{img}}=2$, $T_{\mathrm{ft}}=8$, $L=256$, and $d=768$. To handle asynchronous inputs, we employ timestamp-based windowing where each visual frame captured at 30 Hz is aligned with corresponding F/T samples captured at 200+ Hz within the temporal boundaries defined by consecutive image frames. This preserves high-frequency force dynamics while maintaining correspondence with visual observations.

\textbf{Frequency-aware Multimodal Embedding.} To enable unified processing, we align the encoded modalities using three types of learnable positional embeddings: (i) spatial embeddings, $\mathbf{E}_{\mathrm{spatial}}\in\mathbb{R}^{L\times d}$, which encode within-map positions for visual tokens; (ii) frequency-aware embeddings, $\mathbf{E}_{\mathrm{freq}}\in\mathbb{R}^{T_{\mathrm{ft}}\times d}$, which capture timestamp relationships across modalities with different sampling rates; and (iii) modality embeddings, $\mathbf{E}_{\mathrm{cam1}}, \mathbf{E}_{\mathrm{cam2}}, \mathbf{E}_{\mathrm{ft}}\in\mathbb{R}^{1\times d}$ that encode the source modality of each token. For the hand-eye cameras, frequency-aware embeddings are obtained by resampling $\mathbf{E}_{\mathrm{freq}}$ to $\mathbb{R}^{T_{\mathrm{img}}\times d}$ via linear interpolation \textit{to approximately align with the cameras’ asynchronous timestamps}. These embeddings are added to the visual and F/T tokens and are learned during training, enabling the transformer to model the spatial, frequency, and modality-specific context of each token in the heterogeneous multimodal sequence. As a result, the unified visual tokens $\mathbf{T}'_{\mathrm{cam}}\in\mathbb{R}^{2LT_{\mathrm{img}}\times d}$ and the F/T tokens $\mathbf{T}'_{\mathrm{ft}}\in\mathbb{R}^{T_{\mathrm{ft}}\times d}$ are used as inputs to the bi-directional cross-attention module.

\textbf{Bi-directional Cross-Attention.} The architecture employs bi-directional cross-attention mechanisms to facilitate information exchange between modalities. Visual tokens attend to F/T tokens to incorporate contact dynamics, while F/T tokens attend to visual features to understand spatial context. This design enables learning of rich cross-modal representations that capture interdependencies between visual scenes and physical interactions. Mathematically, the enhanced visual and F/T tokens can be represented as follows:
\begin{gather}
\mathbf{T}^{\prime\prime}_{\mathrm{img}}=
\operatorname{CA_{img\leftarrow ft}}(\mathrm{Q}=\mathbf{T}^{\prime}_{\mathrm{img}},
\mathrm{K}=\mathbf{T}^{\prime}_{\mathrm{ft}},
\mathrm{V}=\mathbf{T}^{\prime}_{\mathrm{ft}})
\\
\mathbf{T}^{\prime\prime}_{\mathrm{ft}}=
\operatorname{CA_{ft\leftarrow img}}(\mathrm{Q}=\mathbf{T}^{\prime}_{\mathrm{ft}},
\mathrm{K}=\mathbf{T}^{\prime}_{\mathrm{img}},
\mathrm{V}=\mathbf{T}^{\prime}_{\mathrm{img}}),
\end{gather}
where $\operatorname{CA}_{a\leftarrow b}$ denotes cross-attention that uses queries from $a$ and keys/values from $b$ (i.e., $\mathrm{Q}=\mathbf{X}_a,\; \mathrm{K}=\mathrm{V}=\mathbf{X}_b$). Finally, we obtain the unified observation representation by concatenating the enhanced visual and F/T tokens, applying a fully connected layer, and then layer normalization:
\begin{equation}
    \mathbf{T}_{\mathrm{obs}}
= \operatorname{LN}\!\left(
  \operatorname{FC}\!\left(
    \operatorname{Cat}\!\bigl(\mathbf{T}^{\prime\prime}_{\mathrm{img}},\, \mathbf{T}^{\prime\prime}_{\mathrm{ft}}\bigr)
  \right)
\right)
\in \mathbb{R}^{(2L T_{\mathrm{img}} + T_{\mathrm{ft}})\times d}.
\end{equation}

\textbf{Transformer Diffusion Policy.} Following the Time-series Diffusion Transformer \cite{chi2023diffusion} architecture, the unified multimodal tokens $\mathbf{T}_{\mathrm{obs}}$ serve as conditioning input for action generation. At each denoising step, noisy action embeddings attend to the multimodal observation features through cross-attention, while maintaining causal self-attention over previous action tokens. The noise prediction network generates action sequences through iterative denoising steps, naturally handling variable-length sequences from different sampling rates without requiring explicit synchronization strategies.

\section{Experimental Results}
We evaluate the proposed \textbf{FMT} on six real-world contact-rich manipulation tasks to validate its ability to fuse high-frequency force and RGB data. 
Section~\ref{subsec:main_table} compares FMT with an RGB-only baseline to quantify the benefits of incorporating high-frequency force information. 
Section~\ref{subsec:ablation} isolates the effects of F/T sensing, frequency-aware multimodal embeddings, and cross-attention via ablation. 
Section~\ref{subsec:ft_freq_comparison} examines how varying the F/T sampling rate from 30–200 Hz affects task performance. 
Finally, Section~\ref{subsec:data_collection_analysis} analyzes data-collection efficiency, comparing ManipForce with direct human demonstrations and teleoperation system.

\subsection{Experimental Setup}
\subsubsection{Evaluation Tasks}
We evaluate our approach on six contact-rich manipulation tasks that demand precise force control and multimodal feedback (Fig.~\ref{fig:Evaluation_Task}). The tasks include \textbf{gear assembly}, which demands precise rotational alignment with controlled axial force, \textbf{LAN plug insertion}, which requires sub-millimeter accuracy and carefully regulated insertion forces to avoid damaging component, and \textbf{box flipping}, a non-prehensile manipulation task requiring coordinated push-and-roll motions guided by contact sensing. We also consider \textbf{open lid}, where the gripper must detect and apply force on a small handle to lift it; \textbf{battery disassembly}, where a tool is inserted into narrow gaps to sense contact force and lift batteries safely; and \textbf{battery insertion}, which involves inserting a spring-loaded battery while maintaining the correct force and direction to prevent slipping or ejection. Together, these tasks evaluate our model’s ability to integrate high-frequency force signals with visual feedback for accurate and robust manipulation. For all tasks, approximately 100 demonstration episodes were collected for training, and each task was evaluated over 20 trials with randomized initial robot and object poses.

\subsubsection{Robot Setup} 
We evaluate our approach on a 7-DOF Franka Panda robot equipped with an AIDIN AFT200 F/T sensor mounted at the wrist, identical to the sensor used in the ManipForce data collection system. 
Tool gravity compensation is applied to the robot’s F/T measurements to ensure consistency with the handheld data collection system. 
A compliance controller is employed to improve contact safety and stability during manipulation. 
The hand-eye cameras are positioned to match the TCP-to-lens distances used in ManipForce, preserving geometric consistency between demonstration collection and robot execution (see Fig.~\ref{fig:intro_examples}).

\begin{figure*}[t]
    \centering
    \includegraphics[width=1\linewidth]{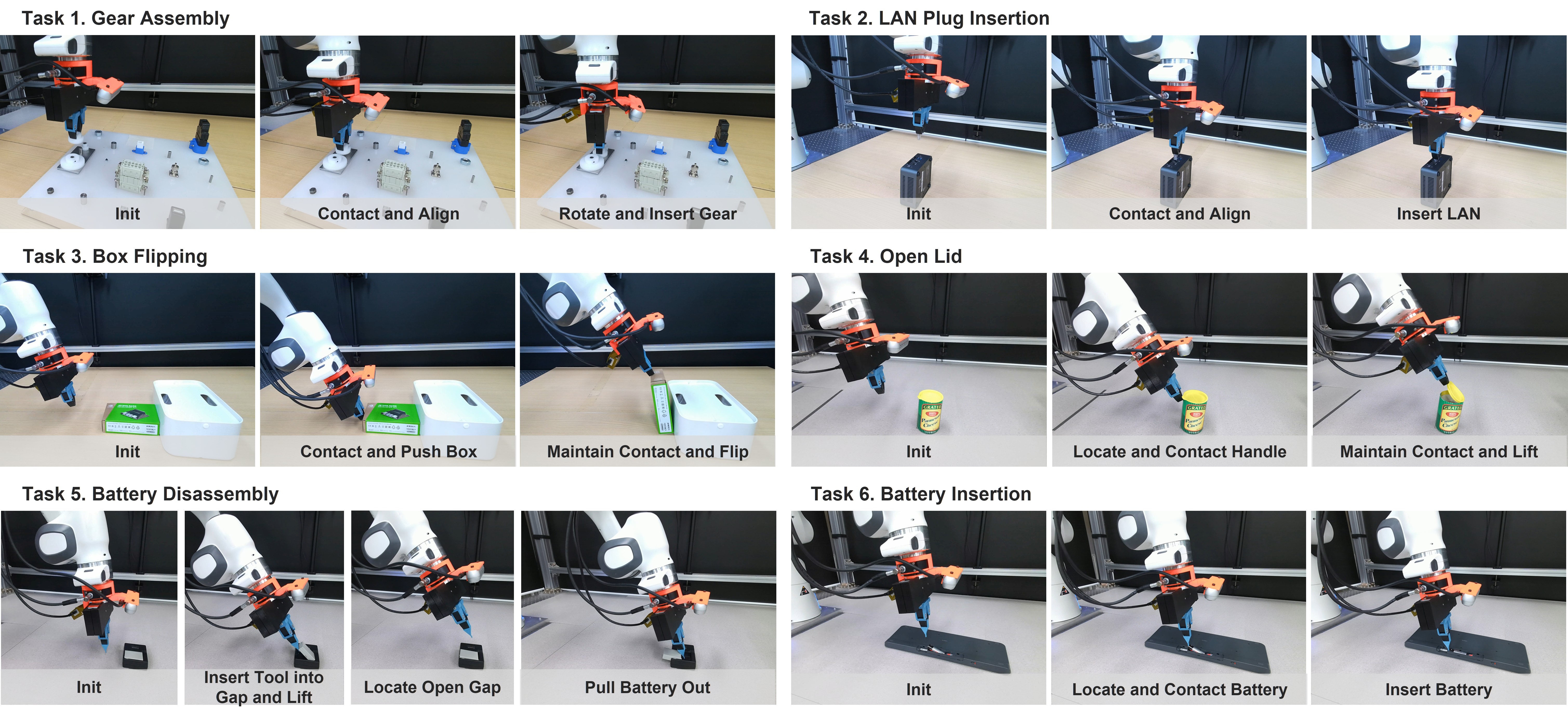}
    \caption{Policy rollout examples from the proposed FMT across the six contact-rich manipulation tasks used for evaluation. Each sequence shows the trained policy executing key stages of each task using integrated RGB–F/T feedback.}
    \label{fig:policy_rollout}
\end{figure*}

\begin{table*}[t]
\centering
\caption{Success rates on contact-rich manipulation tasks comparing our RGB–F/T model with an RGB-only baseline. Each entry reports the proportion of successful trials over 20 evaluation episodes per task.}
\label{tab:comparison}
\resizebox{0.75\textwidth}{!}{
\begin{tabular}{l|ccccccc}
\hline
Method & \makecell{Gear\\Assembly} & \makecell{LAN Plug \\Insertion} & \makecell{Box\\Flipping} & \makecell{Open \\ Lid} & \makecell{Battery\\Disassembly} & \makecell{Battery\\Insertion} \\
\hline
RGB-only\cite{chi2023diffusion} & 0.35 & 0.40 & 0.05 & 0.20 & 0.20 & 0.10 \\
\textbf{FMT} & \textbf{0.95} & \textbf{0.85} & \textbf{0.90} & \textbf{1.00} & \textbf{0.65} & \textbf{0.60} \\
\hline
\end{tabular}
}
\end{table*}

\begin{figure*}[t]
    \centering
    \includegraphics[width=0.75\linewidth]{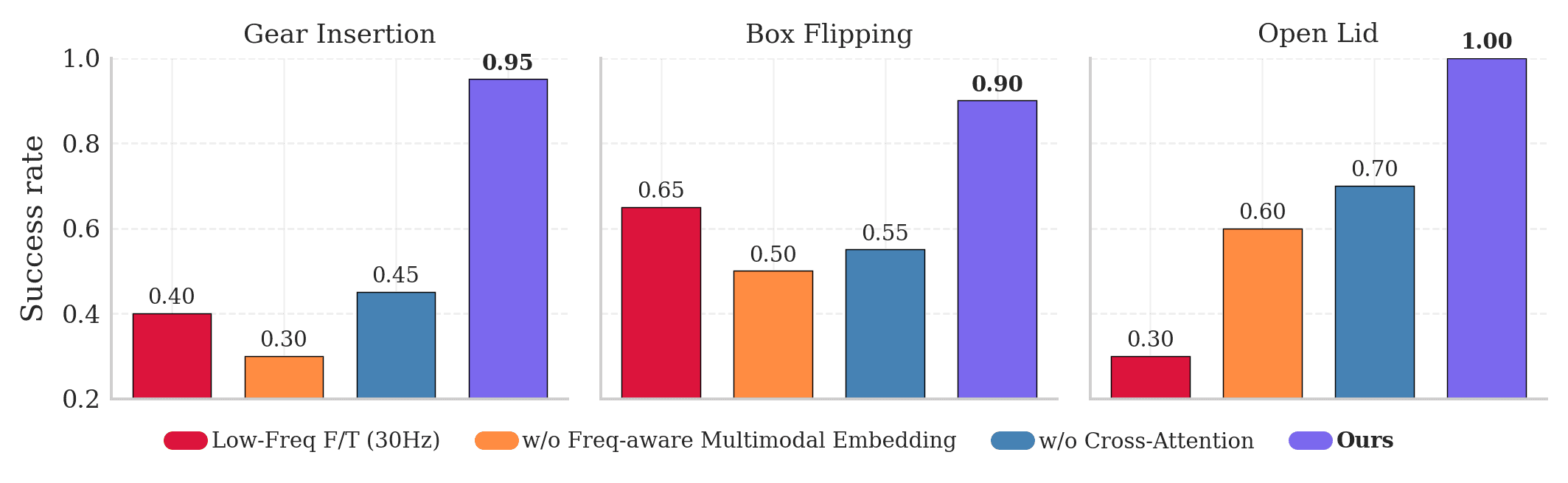}
    \caption{Ablation study on three representative tasks (Gear Assembly, Box Flipping, and Open Lid). 
    We compare the full FMT model with low-frequency F/T input (30~Hz), without positional embeddings, and without cross-attention. 
    FMT consistently outperforms all ablated variants, highlighting the importance of high-frequency F/T sensing, unified positional embeddings, and bi-directional cross-attention for robust manipulation performance.}
    \label{fig:ablation_study}
\end{figure*}

\subsection{Does High-Frequency Force-Aware Policy Learning Improve Performance on Contact-Rich Manipulation?}
\label{subsec:main_table}
We compare two models based on the Transformer Diffusion Policy architecture. 
The \textbf{RGB-only} model~\cite{chi2023diffusion} replaces the original ResNet-18 image encoder with DINOv2-B and uses only RGB tokens as input to the Transformer Diffusion Policy. 
In contrast, our proposed \textbf{FMT} extends the same backbone by incorporating F/T inputs and employing frequency-aware multimodal embeddings with bi-directional cross-attention, enabling fusion of visual and force information at their native sampling rates (Section~\ref{subsec:policy}).
As shown in Table~\ref{tab:comparison}, FMT achieves higher success rates across all six contact-rich tasks, especially in scenarios requiring precise force control or tight clearances. On average, it reaches 83\% success compared to 22\% for the RGB-only model, highlighting the benefit of high-frequency force–aware policy learning.
For example, in \textbf{gear and battery assembly}, which requires repeated fine alignment under rapidly changing forces, FMT maintains precise insertion and stable contact far beyond the RGB-only baseline. 
In \textbf{box flipping}, where sustained contact and continuous force regulation are essential, FMT achieves smooth, reliable manipulation. 
Similarly, in \textbf{open lid}, which hinges on capturing a single transient high-frequency force spike, FMT detects and exploits this cue, achieving robust task execution where the baseline fails. Overall, FMT combines stable contact control with transient event detection, yielding robust and reliable performance across diverse manipulation scenarios (see Fig.~\ref{fig:policy_rollout})

\subsection{How Do FMT Modules Influence Performance Across Contact-Rich Tasks?}
\label{subsec:ablation}

We conducted an ablation study to assess the contribution of FMT’s three core modules: high-frequency force sensing, modal–frequency embeddings, and bi-directional cross-attention. 
We compared the full FMT against three ablated variants:  
\textbf{Low-Freq F/T (30~Hz)} to evaluate the impact of downsampling the force stream;  
\textbf{w/o Freq-aware Multimodal Embeddings} to test the importance of aligning asynchronous RGB and force inputs;  
and \textbf{w/o Cross-Attention} to measure the effect of disabling dynamic vision–force fusion.
Fig.~\ref{fig:ablation_study} shows the success rates of all variants across representative tasks.

Across all six contact-rich tasks, the full FMT achieved the highest success rates, with particularly large gains in tasks demanding fine force modulation or narrow clearances.  
In \textbf{gear assembly}, which requires repeated fine alignment under rapidly changing forces, removing the modal–frequency embeddings caused the steepest performance drop, underscoring their role in precise frequency and cross-modal alignment.  
\textbf{Box flipping}, dominated by slower quasi-static forces, was less sensitive to high-frequency loss but still benefited from both embeddings and cross-attention, showing that even low-frequency tasks gain from improved alignment and fusion.  
By contrast, \textbf{open lid}, which hinges on a single transient high-frequency force spike, exhibited the sharpest degradation under low-frequency input—confirming the need to preserve high-frequency force signatures for critical event detection.

Overall, these results highlight the complementary roles of FMT’s modules.  
High-frequency sensing preserves fine-grained force signatures, modal–frequency embeddings synchronize heterogeneous streams for coherent integration, and bi-directional cross-attention adaptively fuses vision and force for stable, robust policies.  
Together, these elements form the frequency-aware multimodal representations that drive FMT’s strong and consistent performance across diverse contact-rich manipulation tasks.

\subsection{How Does F/T Sampling Frequency Affect Task Performance?}
\label{subsec:ft_freq_comparison}

\begin{figure}[t]
    \centering
    \includegraphics[width=0.8\linewidth]{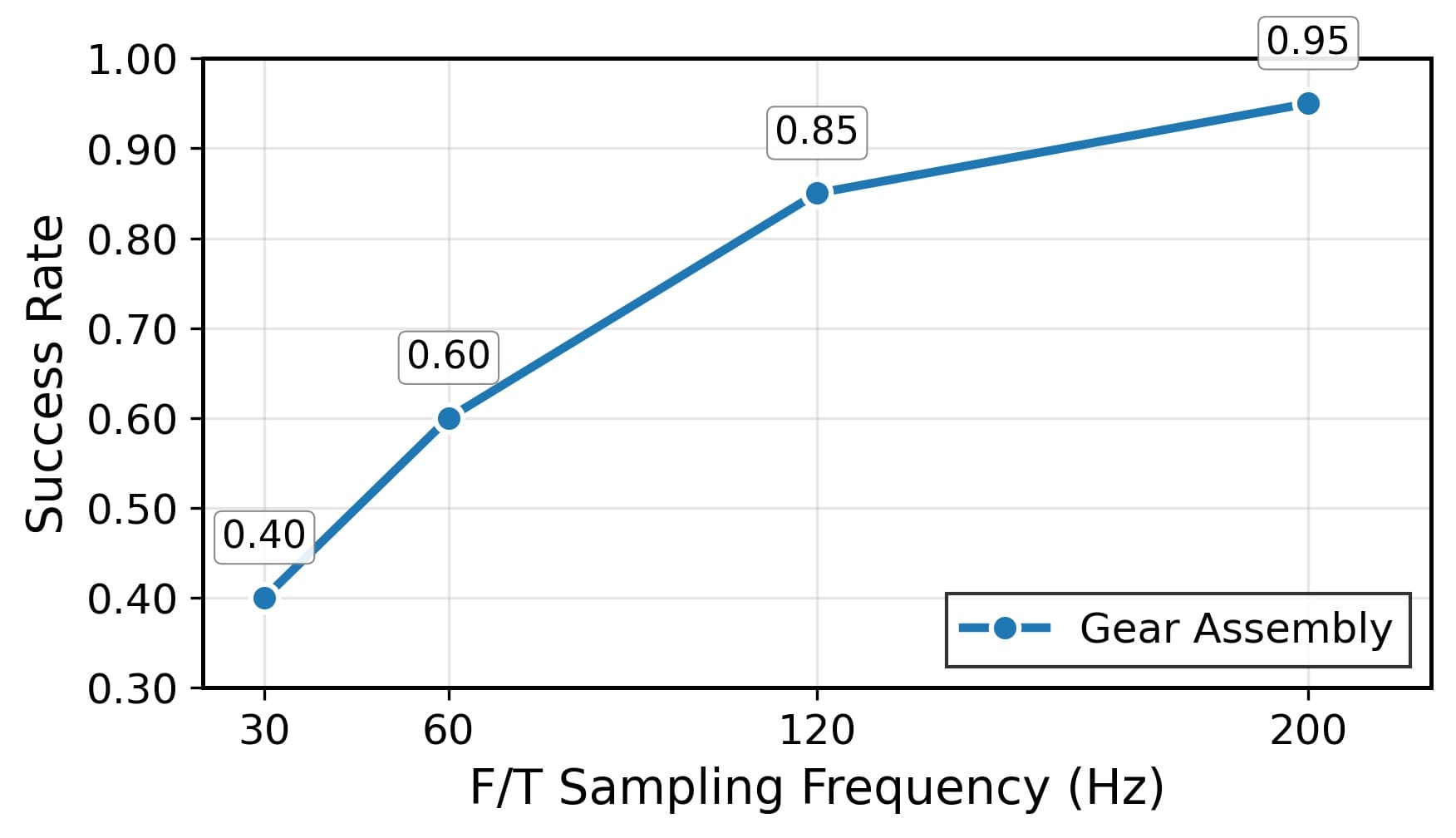}
    \caption{Gear Assembly success rates across F/T sampling frequencies, indicating improved performance at higher frequencies.}
    \label{fig:ft_frequency_performance}
\end{figure}

Fig.~\ref{fig:ft_frequency_performance} presents the success rate of the \textbf{gear assembly} task at different F/T sampling frequencies (30~Hz, 60~Hz, 120~Hz, and 200~Hz). Performance rises monotonically from 0.40 at 30~Hz to 0.95 at 200~Hz, confirming that our model benefits from higher-frequency F/T inputs. At lower sampling rates, the model receives a coarser frequency representation of contact forces, making it harder to detect transient sticking or jamming events and to perform fine rotational adjustments after insertion. In contrast, higher-frequency sensing captures short-duration force spikes and subtle torque changes, enabling more precise corrective actions and smoother alignment. These results demonstrate that high-frequency multimodal sensing directly enhances policy performance in contact-rich assembly tasks and validate our model’s ability to exploit high-bandwidth F/T data.

\subsection{How Effectively Does ManipForce Capture Contact-Rich Demonstrations?}
\label{subsec:data_collection_analysis}
\begin{figure}[t]
    \centering
    \includegraphics[width=\linewidth]{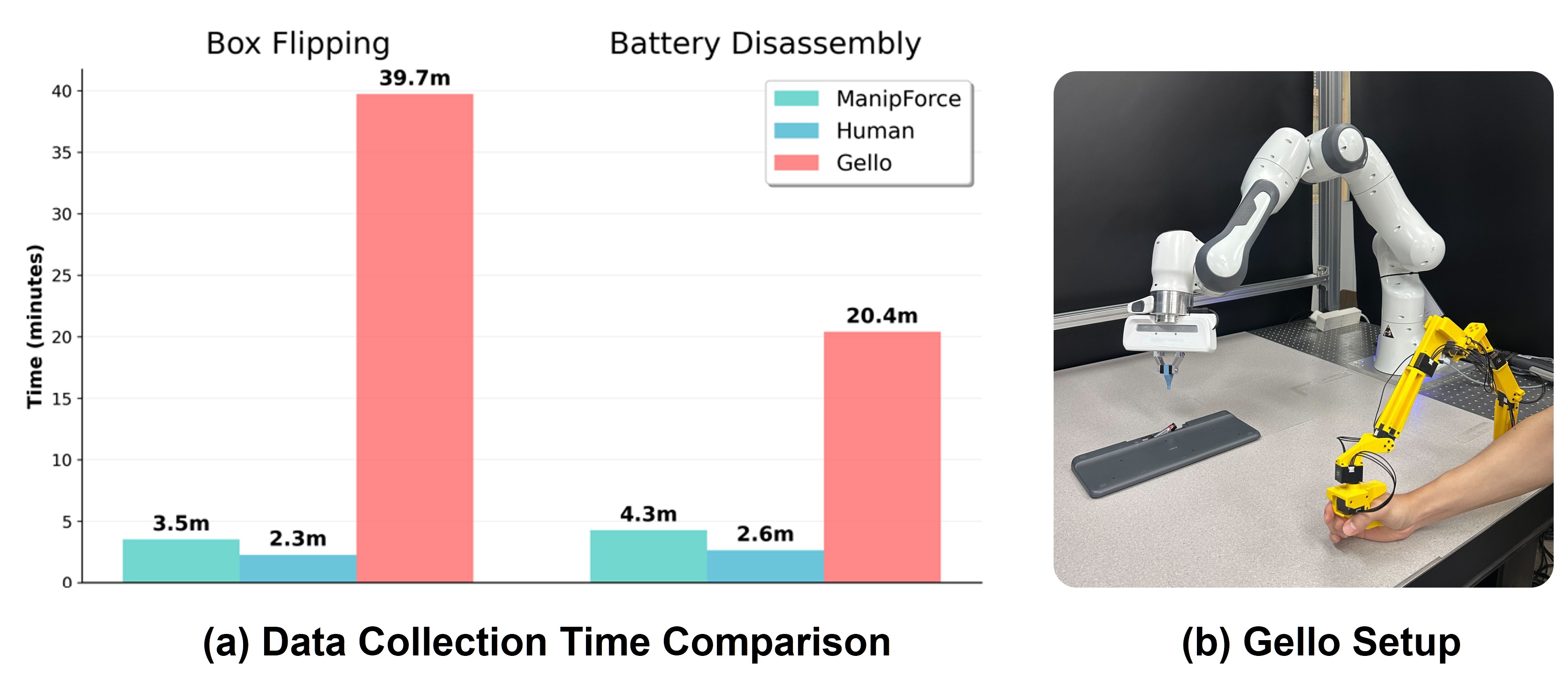}
    \caption{Data collection efficiency comparison among human demonstrations, our ManipForce, and the teleoperation-based Gello, showing ManipForce achieves near-human speed while teleoperation is slower due to the difficulty of maintaining contact and precise force control.}
    \label{fig:data_capturing_comparison}
\end{figure}

We compare data collection efficiency across three settings—\textbf{human demonstrations}, our \textbf{ManipForce} handheld system, and the teleoperation-based \textbf{Gello} system~\cite{wu2024gello} (shown in Fig.~\ref{fig:data_capturing_comparison}b). 
To quantify efficiency, we measured the time required to record demonstrations for two representative tasks—\textbf{box flipping} and \textbf{battery disassembly}—across these settings (results in Fig.~\ref{fig:data_capturing_comparison}a). 
ManipForce achieves efficiency comparable to direct human demonstrations, while Gello requires dramatically longer recording times. 
Beyond efficiency, teleoperation frameworks such as Gello inherently lack haptic feedback, making it difficult for operators to apply accurate interaction forces during contact-rich tasks. 
In contrast, ManipForce leverages natural human operation through a handheld system, ensuring that demonstrations capture both task-relevant motions and the precise interaction forces needed for successful execution.

\section{Conclusion}
We presented ManipForce, a handheld system with dual hand-eye cameras and a wrist-mounted F/T sensor that captures high-frequency multimodal data during natural human demonstrations, enabling direct transfer to robot execution without task-specific calibration. Building on these demonstrations, we introduced FMT, which integrates high-frequency F/T and visual signals for policy learning in contact-rich manipulation, effectively handling their asynchronous sampling rates to achieve precise and stable execution. Across six diverse manipulation tasks, policies trained with FMT achieved an average success rate of about 83\%, substantially outperforming RGB-only baselines and confirming the benefit of high-bandwidth sensing and cross-modal fusion for robust policy learning. While our current study demonstrates strong performance on diverse contact-rich tasks, we plan to extend our framework to longer-horizon and more dexterous tasks requiring more sophisticated gripper behaviors. These directions aim to further enhance the generality and scalability of ManipForce for complex real-world manipulation scenarios.

\section*{Acknowledgments}
\begin{spacing}{0.4}
{\scriptsize This work was supported by the Technology Innovation Program (RS-2024-00442029, Development of Tactile Intelligence in Robotic Hands Based on Tactile Data Learning to Manipulate Irregular Multiple Types of Objects and RS-2024-00423940, Development of Humanoid Robots That Feel Like Humans, Communicate, and Grow through Learning) funded by the Ministry of Trade Industry \& Energy(MOTIE, Korea).}
\end{spacing}

\bibliographystyle{IEEEtran}
\bibliography{references}

\end{document}